\documentclass[10pt,twocolumn,letterpaper]{article}

\usepackage{cvpr}
\usepackage{times}
\usepackage{epsfig}
\usepackage{graphicx}
\usepackage{amsmath}
\usepackage{gensymb}
\usepackage{amssymb}
\usepackage{caption}

\usepackage[pagebackref=true,breaklinks=true,letterpaper=true,colorlinks,bookmarks=false]{hyperref}

\newcommand\blfootnote[1]{%
  \begingroup
  \renewcommand\thefootnote{}\footnote{#1}%
  \addtocounter{footnote}{-1}%
  \endgroup
}

\cvprfinalcopy 

\def\cvprPaperID{1942} 

\newcommand{\env}{\mbox{\sc{RoboTHOR}}}

\ifcvprfinal\fi

\begin{document}

\title{RoboTHOR: An Open Simulation-to-Real Embodied AI Platform}

\author{Matt Deitke$^{2*}$, Winson Han$^{1*}$, Alvaro Herrasti$^{1*}$,
              Aniruddha Kembhavi$^{1,2*}$, Eric Kolve$^{1*}$, \\ Roozbeh Mottaghi$^{1,2*}$,
              Jordi Salvador$^{1*}$, Dustin Schwenk$^{1*}$, Eli VanderBilt$^{1*}$, 
              Matthew Wallingford$^{2*}$, \\ Luca Weihs$^{1*}$, Mark Yatskar$^{1*}$,  
              Ali Farhadi$^{2}$ \\
              $^1$PRIOR @ Allen Institute for AI\,\,\,\,\, $^2$University of Washington \\
              \url{ai2thor.allenai.org/robothor}
}

\twocolumn[{
\renewcommand\twocolumn[1][]{#1}
\maketitle
\vspace*{-0.5cm}
\centering
\includegraphics[width=.99\linewidth]{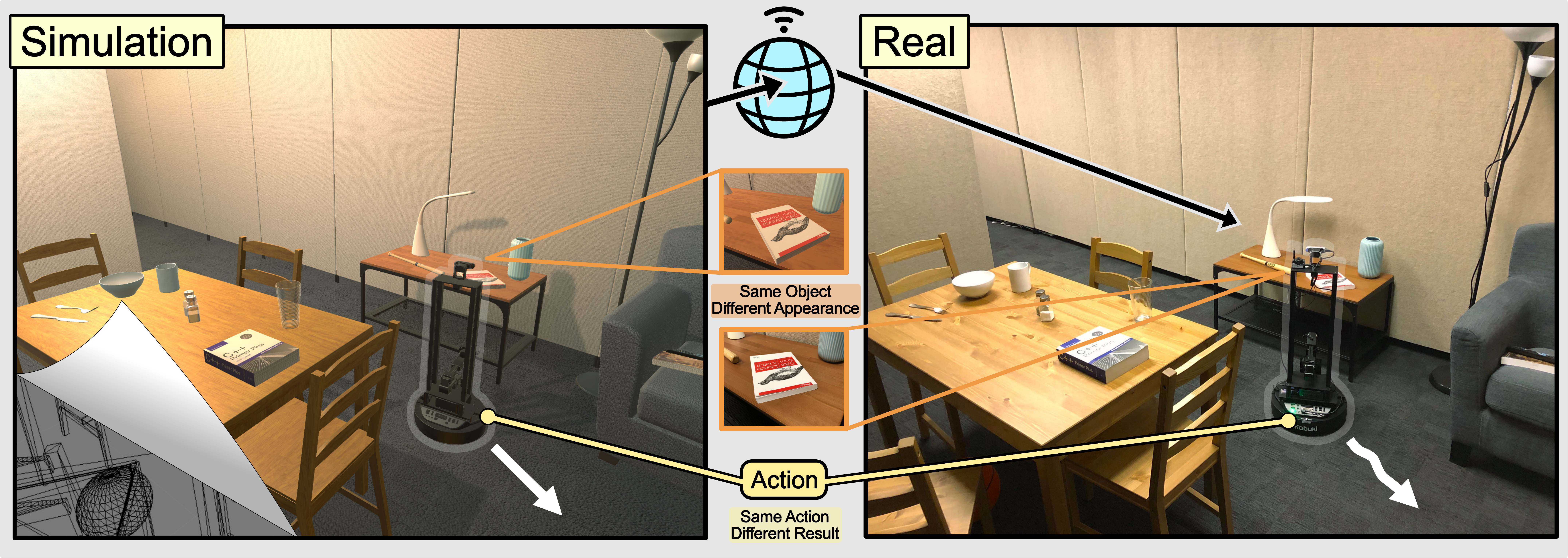}
\vspace{-.2cm}
\captionof{figure}{
We present \env, a platform to develop and test embodied AI agents with corresponding environments in simulation and the physical world. The complexity of environments in \env\ along with disparities in appearance and control dynamics between simulation and reality pose new challenges and open many avenues for further research.
}
\label{fig:teaser}
\vspace*{0.3cm}
}]

\maketitle
\thispagestyle{empty}

\begin{abstract}

\blfootnote{$*$ Alphabetically listed equal contribution}Visual recognition ecosystems (e.g.\ ImageNet, Pascal, COCO) have undeniably played a prevailing role in the evolution of modern computer vision. We argue that interactive and embodied visual AI has reached a stage of development similar to visual recognition prior to the advent of these ecosystems. Recently, various synthetic environments have been introduced to facilitate research in embodied AI. Notwithstanding this progress, the crucial question of how well models trained in simulation generalize to reality has remained largely unanswered.  
The creation of a comparable ecosystem for simulation-to-real embodied AI presents many challenges: (1) the inherently interactive nature of the problem, (2) the need for tight alignments between real and simulated worlds, (3) the difficulty of replicating physical conditions for repeatable experiments, (4) and the associated cost. In this paper, we introduce \env\ to democratize research in interactive and embodied visual AI. \env\ offers a framework of simulated environments paired with physical counterparts to systematically explore and overcome the challenges of simulation-to-real transfer, and a platform where researchers across the globe can remotely test their embodied models in the physical world. As a first benchmark, our experiments show there exists a significant gap between the performance of models trained in simulation when they are tested in both simulations and their carefully constructed physical analogs. We hope that \env\ will spur the next stage of evolution in embodied computer vision. 

\end{abstract}

\section{Introduction}

For decades, the AI community has sought to create perceptive, communicative and collaborative agents that can augment human capabilities in real world tasks. While the advent of deep learning has led to remarkable breakthroughs in computer vision~\cite{krizhevsky2012imagenet, He_2016_CVPR, faster_rcnn} and natural language processing~\cite{Peters2018DeepCW, Devlin2019BERTPO}, creating active and intelligent embodied agents continues to be immensely challenging.

The widespread availability of large and open, computer vision and natural language datasets~\cite{Russakovsky2014ImageNetLS, Lin2014MicrosoftCC, Rajpurkar2016SQuAD10, Wang2018GLUEAM}, massive amounts of compute, and standardized benchmarks have been critical to this fast progress. In stark contrast, the considerable costs involved in acquiring physical robots and experimental environments, compounded by the lack of standardized benchmarks are proving to be principal hindrances towards progress in embodied AI. In addition, current state of the art supervised and reinforcement learning algorithms are data and time inefficient; impeding the training of embodied agents in the real world.

Recently, the vision community has leveraged progress in computer graphics and created a host of simulated perceptual environments such as AI2-THOR~\cite{ai2thor}, Gibson~\cite{gibson}, MINOS~\cite{minos} and Habitat~\cite{habitat}, with the promise of training models in simulation that can be deployed on robots in the physical world. These environments are free to use, continue to be improved and lower the barrier of entry to research in real world embodied AI; democratizing research in this direction. This has led to progress on a variety of tasks in simulation, including visual navigation \cite{gupta17,wortsman19}, instruction following \cite{anderson18,Wang_2018_ECCV} and embodied question answering \cite{gordon18,das18}. But the elephant in the room remains: \emph{How well do these models trained in simulation generalize to the real world?}

While progress has been ongoing, the large costs involved in undertaking this research has restricted pursuits in this direction to a small group of well resourced organizations. We believe that creating a free and accessible framework that pairs agents acting in simulated environments with robotic counterparts acting in the physical world will open up this important research topic to all\textemdash bringing faster progress and potential breakthroughs. As a step towards this goal, we present \env.

\env\ is a platform to develop artificial embodied agents in simulated environments and test them in both, simulation as well as the real world. A key promise of \env\ is to serve as an open and accessible benchmarking platform to stimulate reproducible research in embodied AI. With this in mind, it has been designed with the following properties:

\begin{itemize}
    \setlength\itemsep{0.01em}
    \item \textbf{Simulation and Real Counterparts} - \env\ consists of a training and validation corpus of 75 scenes in simulation at present. A total of 14 test-dev and test-standard scenes are present in simulation and their counterparts constructed in the physical world. Scenes are designed to have diverse wall and furniture layouts, and all are densely populated by a variety of object categories. Figure~\ref{fig:teaser} shows a view of the dining room in one of the test-dev scenes, in simulation as well as real.
    \item \textbf{Modular} - Scenes in \env\ are built in a modular fashion, drawing from an asset library containing wall structures, flooring components, ceilings, lighting elements, furniture pieces and objects; altogether totaling 731 unique assets distributed across scenes. This enables scene augmentation and easy expansion of \env\ to fit the needs of researchers.
    \item \textbf{Re-configurable} - The physical environments are also built using modular and movable components, allowing us to host scenes with vastly different layouts and furniture arrangements within a single physical space. This allows us to scale our test corpora while limiting their cost and physical footprint. Reconfiguring the space to a new scene can be accomplished in roughly 30 minutes.
    \item \textbf{Accessible to all} - The simulation environment, assets and algorithms we develop will be open source. More critically, researchers from all over the world will be able to remotely deploy their models on our hardware at no cost to them. We will be setting up a systematic means of reserving time in our environment.
    \item \textbf{Replicable} - The physical space has been designed to be easily replicable by other researchers should they wish to construct their own physical environment. This is achieved through open sourced plans, readily available building components, IKEA furniture, and a low cost amounting to roughly \$10,000 in materials and assets to create the physical space. In addition, we use LoCoBot, an inexpensive and easily obtainable robot.
    \item \textbf{Benchmarked} - In addition to open sourcing baseline models, we will host challenges involving several embodied AI tasks with a focus on the ability to transfer these models successfully onto robots running in a physical environment.
\end{itemize}

\env\ has been designed to support a variety of embodied AI tasks. In this work we benchmark models for semantic navigation, the task of navigating to an instance of the specified category in the environment. The complexity and density of scenes in \env\ renders this task quite challenging, with humans requiring 49.5 steps (median statistic) to find the target object. We train a set of competitive models using a pure reinforcement learning (RL) approach with asynchronous actor critic (A3C) on the simulated training environments, measure their performance on the simulated as well as real validation environments and arrive at the following revealing findings. (1) Similar to findings in past works such as \cite{wortsman19}, semantic navigation models struggle with generalizing to unseen environments in simulation. We show that their performance takes an even larger hit when deployed onto a physical robot in the real world. (2) We analyze simulated and real world egocentric views and find a disparity in feature space in spite of the images from the two modalities looking fairly similar to the naked eye; a key factor affecting the transfer of policies to the real world. (3) As expected and noted in previous works, control dynamics in the real world vary significantly owing to motor noise, slippage, and collisions. (4) Off the shelf image correction mechanisms such as image-to-image translation do not improve performance.

These findings reveal that training embodied AI models that generalize to unseen simulated environments and further yet to the real world remains a daunting challenge; but also open up exciting research frontiers. We hope that \env\ will allow more research teams from across the globe to participate in this research  which will result in new model architectures and learning paradigms that can only benefit the field.
\section{Related Work}
\noindent \textbf{Embodied AI Environments.} In recent years, several synthetic frameworks have been proposed to investigate tasks including visual navigation, task completion and question answering in indoor scenes \cite{ai2thor,minos,house3d,home,chalet,gibson,VirtualHome,habitat}. These free virtual environments provide excellent testbeds for embodied AI research by abstracting away the noise in low-level control, manipulation and appearance and allowing models to focus on the high-level end goal. \env\ provides a framework for studying these problems as well as for addressing the next frontier: transferring models from simulation to the real world.

Robotics research platforms~\cite{baxter, sawyer} have traditionally been expensive to acquire. More recent efforts have led to low cost robot solutions~\cite{srinivasa2019mushr, mitracecar, pyrobot2019} opening up the space to more research entities.
There has been a long history of using simulators in conjunction with physical robots. These largely address tasks such as object manipulation using robotic arms~\cite{chebotar18} and autonomous vehicles~\cite{srinivasa2019mushr, airsim}.

\noindent \textbf{Visual Navigation.} In this paper, we explore models for the task of visual navigation, a popular topic in the robotics and computer vision communities. The navigation problem can be divided into two broad categories, spatial navigation and semantic navigation. Spatial navigation approaches \cite{thorpe88,fox99,urmson08,hadsell09,henry10,shen2011,zhu17,richter17,kahn17,chen18,chaplot19} typically address navigating towards a pre-specified coordinate or a frame of a scene and they focus on understanding the geometry of the scene and learning better exploration strategies. For example, \cite{zhu17} address navigation towards a given input image, \cite{chaplot19} address navigation towards a point in a scene and \cite{kahn17} learn a collision-free navigation policy. Semantic navigation approaches \cite{gupta17,yang19,wortsman19,wu19,mirowski17,savinov18,MousavianArxiv18} attempt to learn the semantics of the target in conjunction with navigation. For example, \cite{yang19} use prior knowledge of object relations to learn a policy that better generalize to unseen scenes or objects. \cite{wortsman19} use meta-learning to learn a self-supervised navigation policy toward a specified object category. \cite{wu19} use prior knowledge of scene layouts to navigate to a specific type of room. We benchmark models on the task of semantic navigation.

Navigation using language instructions has been explored by \cite{anderson18,Wang_2018_ECCV,fried2018,ke2019,wang2019,ma2019a,ma2019b}. This line of work has primarily been tested in simulation; transferability to the real world remains an open question and addressing this via \env\ is a promising future endeavour. Navigation has also been explored in other contexts such as autonomous driving (e.g., \cite{chen15deepdrive,xu17}) or city navigation (e.g., \cite{mirowski18,chen19}). In this work, we focus on indoor navigation.

\noindent \textbf{Sim2Real Transfer.} Domain adaptation in general as well as Sim2Real in particular, have a long history in computer vision. There are different techniques to adapt models from a source domain to a target domain. The main approaches are based on randomization of the source domain to better generalize to the target domain \cite{tobin17,james17,sadeghi17,peng18,tan18}, learning the mapping between some abstraction or higher order statistics of the source and target domains \cite{hoffman16,sun16,ganin16,long15,zhang17,mueller18,zhu18}, interpolating between the source and the target domain on a learned manifold \cite{gong12,caseiro15}, or generating the target domain using generative adversarial training \cite{bousmalis16,taigman17,shrivastava17,bousmalis18,hoffman18,huang18}. \env\ enables source randomization via scene diversity and asset diversity. We also experiment with using an off the shelf target domain mapping method, the GAN-based model of \cite{CycleGAN2017}.

\section{RoboTHOR}

\begin{figure*}[tp]
    \centering
    \includegraphics[width=40pc]{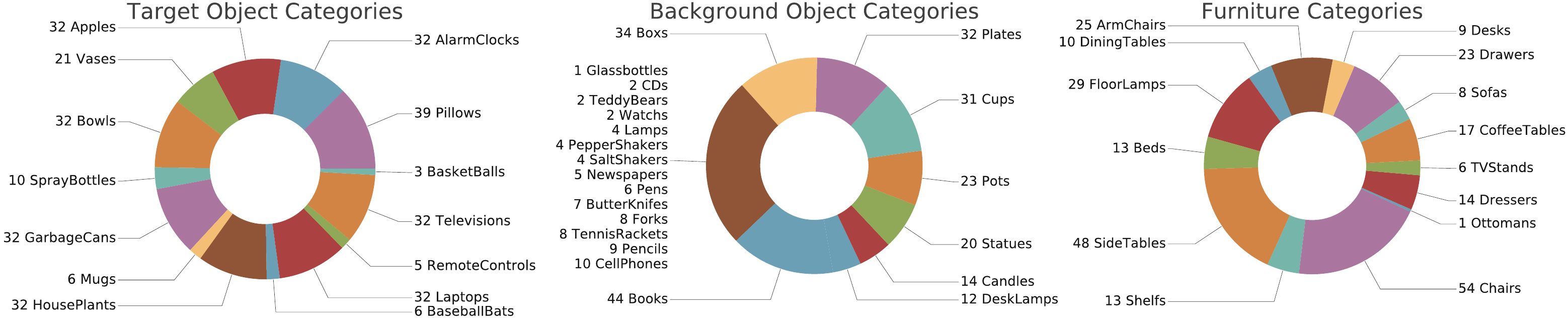}
    \caption{Distribution of object categories in \env\ }
    \label{fig:piechart} 
\end{figure*}

State of the art learning algorithms for embodied AI use reinforcement learning based approaches to train models, which typically require millions of iterations to converge to a reasonable policy. Training policies in the real world with real robots would take years to complete, due to the mechanical constraints of robots. Synthetic environments, on the other hand, provide a suitable platform for such training strategies, but how well models trained in simulation transfer to the real world, remains an open question. \env\ is a platform, built upon the AI2-THOR framework \cite{ai2thor} to build and test embodied agents with an emphasis on studying this problem of domain transfer from simulation to the real world. 

\begin{figure}[h]
    \centering
    \includegraphics[width=20pc]{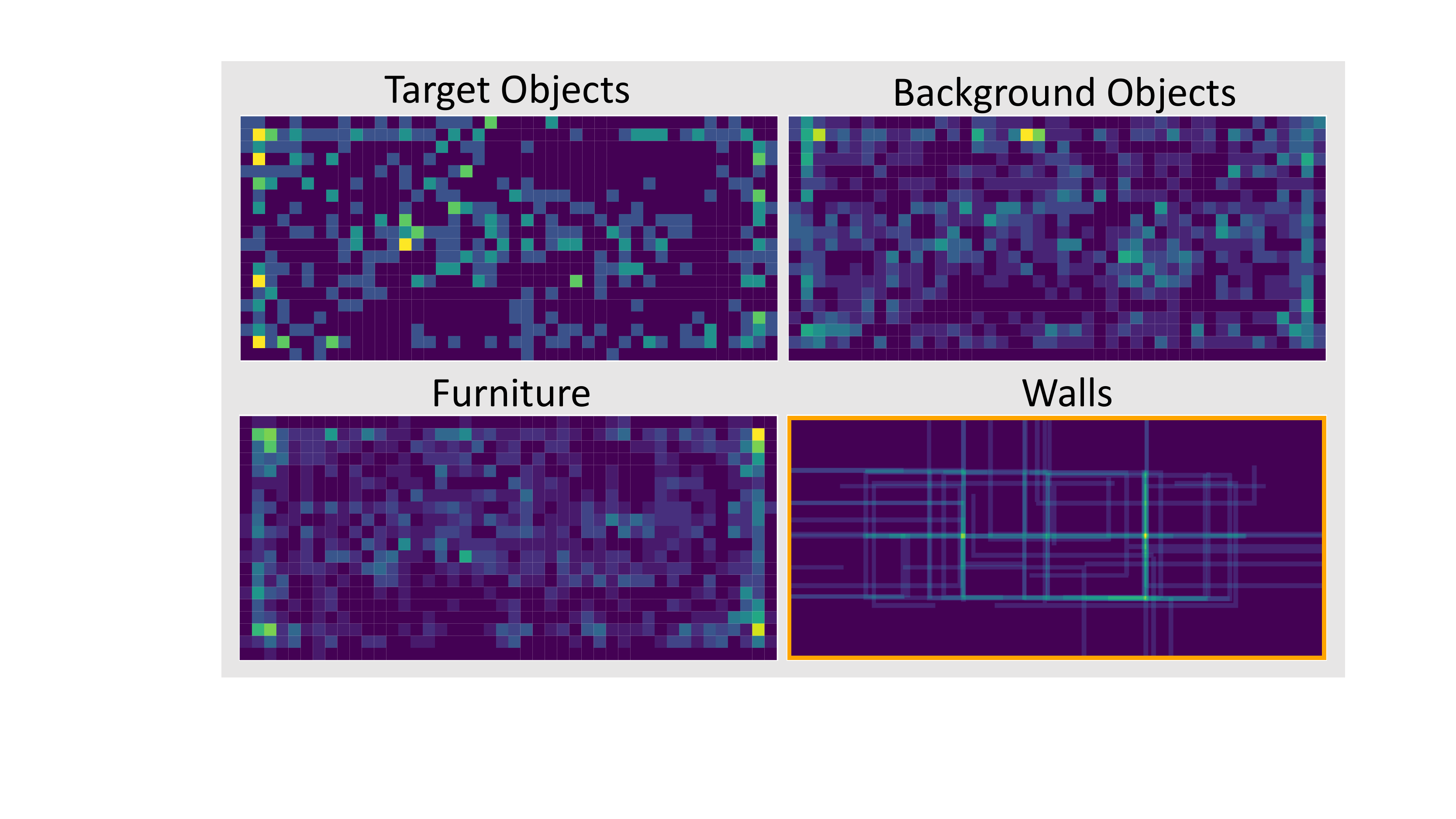}
    \caption{\textbf{Spatial distribution of objects and walls.} Heatmaps illustrate the diverse spatial distribution of target objects, background objects, furniture, and walls. }
    \label{fig:obj_wall_dist} 
\end{figure}

\noindent \textbf{Scenes.} \env\ consists of a set of 89 apartments, 75 in train/val (we use 60 for training and 15 for validation), 4 in test-dev (which are used for validation in the real world) and 10 in test-standard (blind physical test set) drawn from a set of 15, 2 and 5 wall layouts respectively. Apartments that share the same wall layout have completely different room assignments and furniture placements (for example, a bedroom in one apartment might be an office in another). Apartment layouts were designed to encompass a wide variety of realistic living spaces. Figure~\ref{fig:obj_wall_dist} shows a heatmap of wall placements across the train/val subset. This set of apartments is only instantiated in simulation, while the test-dev and test-standard apartments are also substantiated in the physical world. The layouts, furniture, objects, lighting, etc. of the simulation environments have been designed carefully so as to closely resemble the corresponding scenes in their physical counterparts, while avoiding any overlap between the wall layouts and object instances among train/val, test-dev and test-standard. This resemblance will enable researchers to study the discrepancies between the two modalities and systematically identify the challenges of the domain transfer. 

\begin{figure}[h]
    \centering
    \includegraphics[width=18pc]{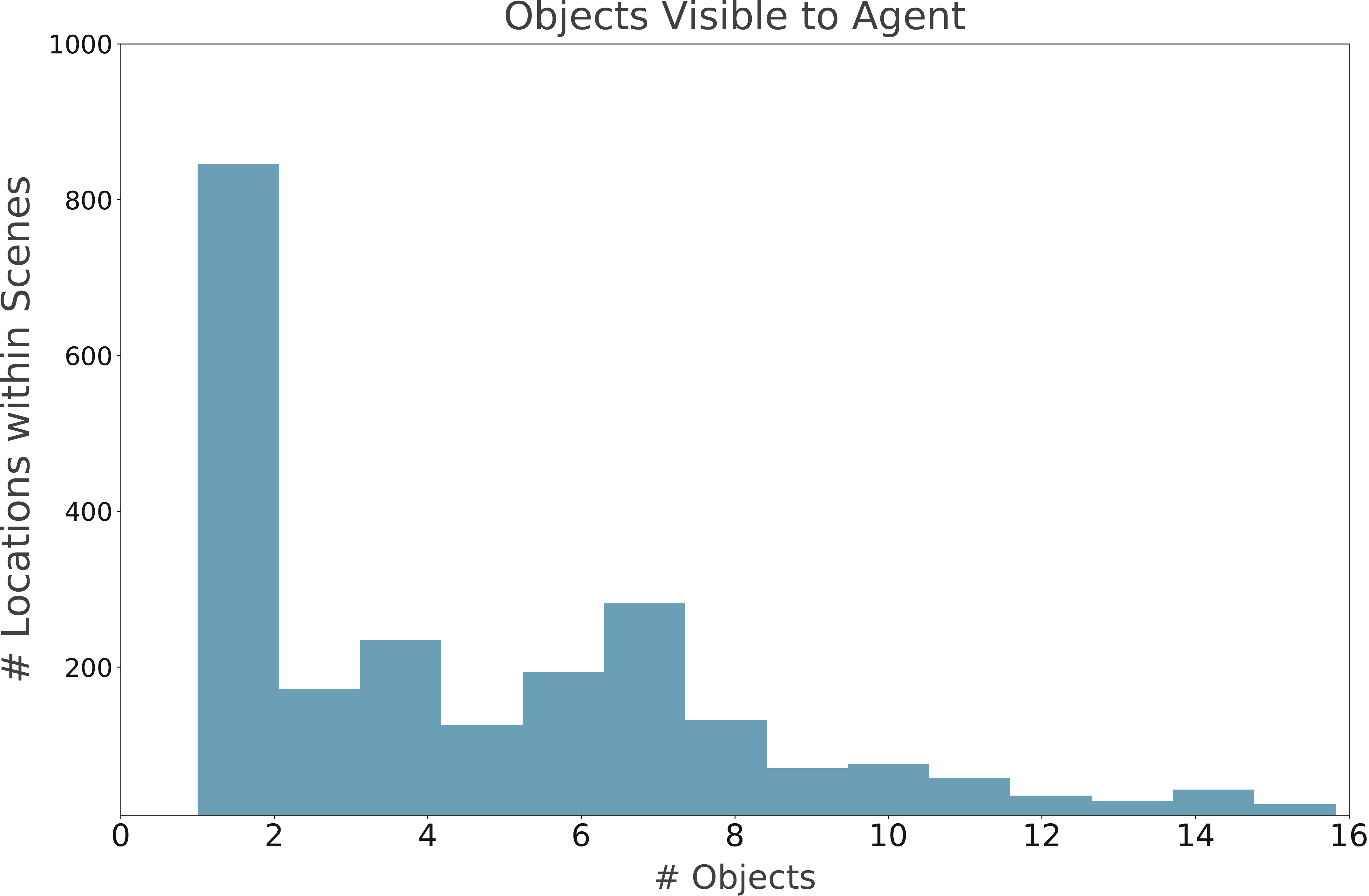}
    \caption{\textbf{Object visibility statistics.} The distribution of objects visible to an agent at a single time instant.}
    \label{fig:val_objects_vis}
    \vspace{-0.5cm}
\end{figure}

\noindent \textbf{Assets.} A guiding design principle of \env\ is \emph{modularity}, which allows us to easily augment and scale scenes. 
A large asset library was created by digital artists from which scenes were created by selectively drawing from these assets. This is in contrast to environments that are based on 3D scans of rooms which are challenging to alter and interact with. The framework includes 11 types of furniture (e.g. TV stands and dining tables) and 32 types of small objects (e.g. mugs and laptops) across all scenes. The majority of real furniture and objects were gathered from IKEA. Among the small objects categories, 14 are designated as targets and guaranteed to be found in all scenes for use in semantic navigation tasks. In total there are 731 unique object instances in the asset library with no overlap among train/val, test-dev and test-standard scenes. Figure~\ref{fig:piechart} shows the distribution of object categories amongst the asset library. We distribute object categories as uniformly as possible to avoid bias toward specific locations. Figure~\ref{fig:obj_wall_dist} shows the spatial distribution of target objects, background objects and furniture in the scenes. Figure~\ref{fig:val_objects_vis} shows the distribution of the number of visible objects in a single frame. A large number of frames consist of the agent looking at the wall as is common in apartments, but outside these views many objects are visible to the agent at any given point as it navigates the environment.

\noindent \textbf{Physical space.} The physical space for \env\ is $8.8m \times 3.9m$. The space is partitioned into rooms and corridors using ProPanel walls, which are designed to be lightweight and easy to set up and tear down, allowing us to easily configure a new apartment layout in a few minutes.

\noindent \textbf{Agent.} The physical robot used is a LoCoBot\footnote{\url{http://www.locobot.org/}}, which is equipped with an Intel RealSense RGB-D camera. We replicate the robot in simulation with the same physical and camera properties. To mimic the noisy dynamics of the robot movements, we add noise to the controller in simulation. The noise parameters are estimated by manually measuring the error in orientation and displacement over multiple runs.

\noindent \textbf{API.} To enable a seamless switch between the synthetic and the real environments, we provide an API that is agnostic to the underlying platform. Hence, agents trained in simulation can be easily deployed onto the LoCoBot for testing. The API was built upon the PyRobot~\cite{pyrobot2019} framework to manage control of the LoCoBot base as well as camera.

\noindent \textbf{Connectivity.} A main goal of this framework is to provide access to researchers across the globe to deploy their models onto this physical environment. With this in mind, we developed the infrastructure for connecting to the physical robot or the simulated agent via HTTP. A scheduler prevents accessing the same physical hardware by multiple parties. 

\noindent \textbf{Localization.} We installed Super-NIA-3D localization modules across the physical environment to estimate the location of the robot and return that to the users. For our experiments, we do not use the location information for training as this type of training signal is not usually available in real world scenes. We use this location information only for evaluation and visualization.

\section{Visual Semantic Navigation}
In this paper, we benchmark models for the task of visual semantic navigation, i.e. navigating towards an instance of a pre-specified category. \env\ enables various embodied tasks such as question answering, task completion and instruction following. Navigation is a key component of all these tasks and is a necessary and important first step towards studying transfer in the context of embodied tasks.

Visual semantic navigation evaluates the agent's capabilities not only in avoiding obstacles and making the right moves towards the target, but also understanding the semantics of the scene and targets. The agent should learn how different instances of an object category look like and should be able to reason about occlusion, scale changes and other variations in object appearance. 

More specifically, our goal is to navigate towards an instance of an object category specified by a noun (e.g., Apple) given ego-centric sensory inputs. The sensory input can be an RGB image, a depth image, or combination of both. At each time step the agent must issue one of the following actions: \emph{Move Ahead}, \emph{Rotate Right}, \emph{Rotate Left}, \emph{Look Up}, \emph{Look Down}, \emph{Done}. The action \emph{Done} signifies that the agent reports that it has reached its goal and leads to an end of episode. We consider an episode successful if (a) the object is in view (b) the agent is within a threshold of distance to the target and (c) the agent reports that it observes the object. The starting location of the agent is a random location in the scene.

The motion of the agent in the simulated world is stochastic in nature, mirroring its behavior in the real world. This renders the task more challenging. Previous works such as \cite{wortsman19} consider agent motion along the axes on a grid. But given the end goal of navigating in the real world with motor noise and wheel slippage, deterministic movements in training lead to sub optimal performance during testing.

\begin{figure}[h]
    \centering
    \includegraphics[width=18pc]{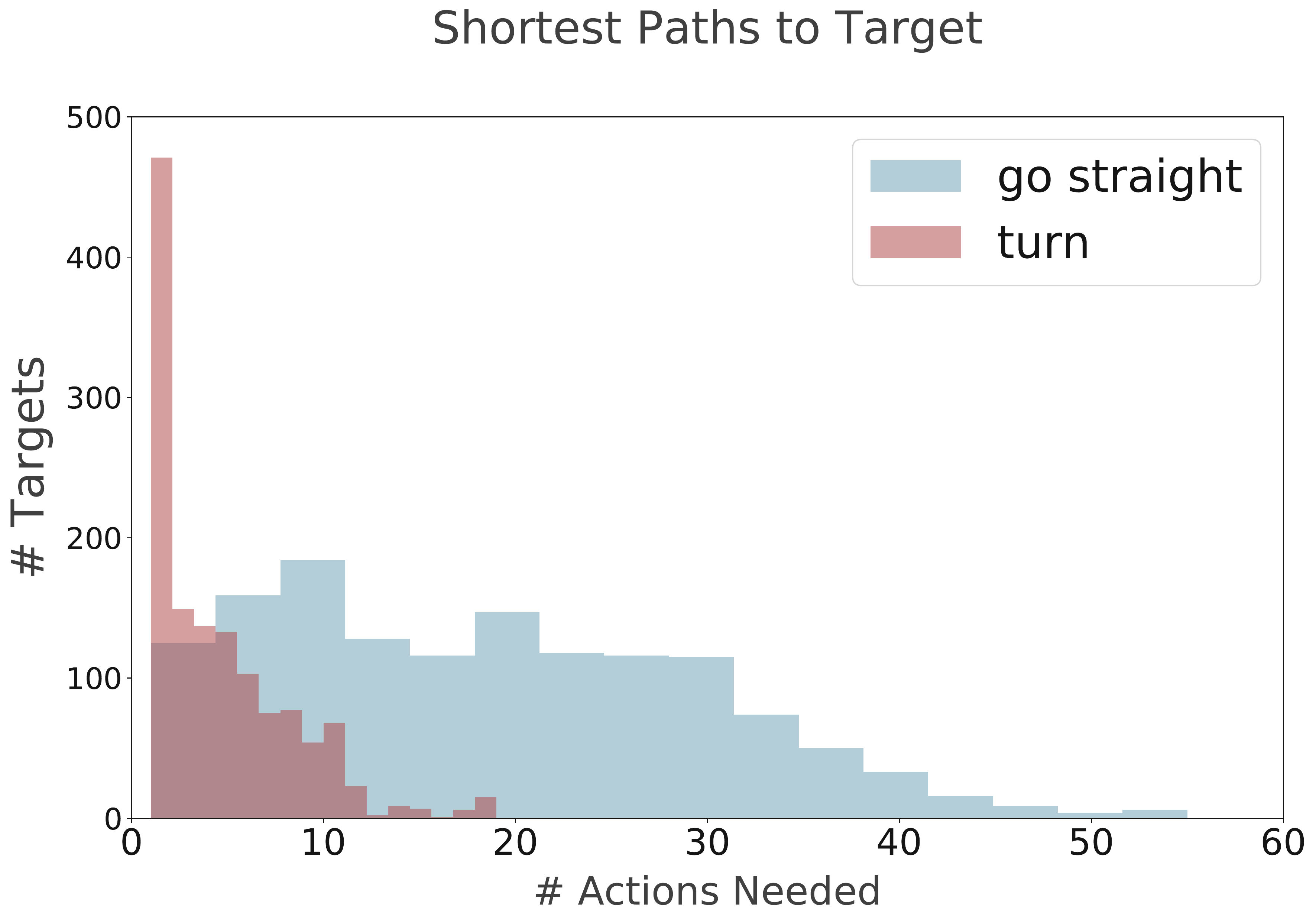}
    \caption{\textbf{Histogram of actions along the shortest path.} The number of actions invoked along the shortest paths to targets in the training scenes. Note that the shortest path is very difficult to obtain in practice, since it assumes \emph{a priori} knowledge of the scene.}
    \label{fig:shortest_path_actions_needed} 
    \vspace{-0.2cm}
\end{figure}

\begin{figure}[h]
    \centering
    \includegraphics[width=18pc]{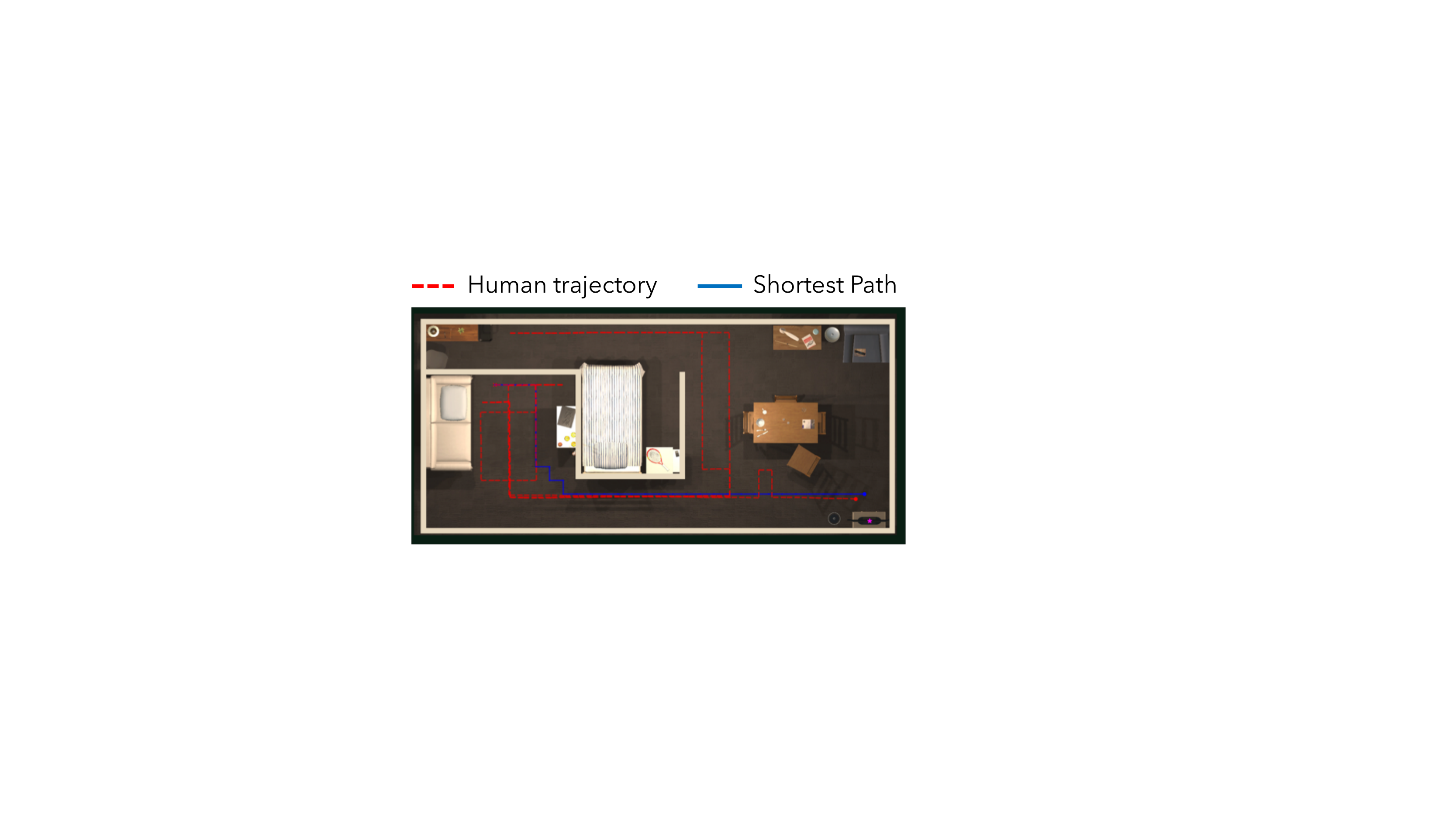}
    \caption{\textbf{Example human trajectory.} The shortest path to a target vs the path taken by a human from the same starting location are visualized. The human wanders around looking for the TV and on seeing it, walks straight towards it.}
    \label{fig:human_path} 
    \vspace{-0.3cm}
\end{figure}

The semantic navigation task is very challenging owing to the size and complexity of the scenes. Figure~\ref{fig:shortest_path_actions_needed} shows the lengths of shortest paths to the target objects, in terms of the \emph{Move Ahead} and \emph{Rotate} actions. But shortest path statistics are a bit misleading for the task of semantic navigation because they assume that the agent already knows the location it must travel towards. In fact, the agent must explore until it observes the target, and then move swiftly towards it. We conducted a study where humans were posed with the problem of navigating in scenes in \env\ (simulation) to find targets. The median number of steps was 49.5 (compared to 22.0 for shortest paths), illustrating the exploration nature of the task. Figure~\ref{fig:human_path} shows an example trajectory from a human compared to the corresponding shortest path.

\subsection{Baseline models}
We measure the performance of the following baseline models:

\noindent \textbf{Random} - This model chooses an action randomly amongst the set of possible actions. This includes invoking the \emph{Done} action. The purpose of this baseline is to ascertain if the scenes and starting locations are not overly simplistic.

\noindent \textbf{Instant Done} - This model invokes the \emph{Done} action at the very first time step. The purpose of this baseline is to measure the percentage of trivial starting locations.

\noindent \textbf{Blind} - This model receives no sensory input. It only consists of an LSTM with simply a target embedding input. Its purpose is to establish a baseline that can only leverage starting and target location bias in the dataset.

\noindent \textbf{Image-A3C} - The agent perceives the scene at time $t$ in the form of an image $o_t$. The image is fed into a pre-trained and frozen ResNet-18 to obtain a $7 \times 7 \times 512$ tensor, followed by two $1 \times 1$ convolution layers to reduce the channel depth to 32 and finally concatenated with a 64 dimensional embedding of the target word and provided to an LSTM which generates the policy. The agent is trained using the asynchronous advantage actor-critic  (A3C)~\cite{Mnih2016AsynchronousMF} formulation, to act to maximize its expected discounted cumulative reward. Two kinds of rewards are provided: a positive success reward and a negative step reward to encourage efficient paths.

\noindent \textbf{Image+Detection-A3C} - The image is fed into Faster-RCNN~\cite{faster_rcnn} trained on the MSCOCO dataset. Each resulting detection has a category, probability and box locations and dimensions, which are converted to an embedding using an MLP. The resulting set of embeddings are converted into a tensor $7 \times 7 \times 120$ by aligning detection boxes to spatial grid cells and only considering the top 3 boxes per location. This tensor is concatenated to the tensor obtained for the image modality and then processed as above.

\subsection{Metrics}
We report Success Rate and Success weighted by Path Length (SPL)~\cite{spl}, common metrics for evaluating navigation models. In addition, we also report path lengths in terms of the number of actions taken as well as distance travelled. This quantifies the exploration carried out by the agent.
\section{Experiments}

\noindent \textbf{Training.} We train models to navigate towards a single target, a Television with an action space of 4 including \emph{Move Ahead}, \emph{Rotate Right}, \emph{Rotate Left} and \emph{Done}. The rotation actions rotate the agent by 45 degrees. The agent must issue the \emph{Done} action when it observes the object. If an agent is successful, it receives a reward of +5. The step penalty at each time step is -0.01. To mimic the noisy dynamics of the real robot, we add noise to the movement of the virtual agent. For translation, we add a gaussian noise with mean 0.001m and standard deviation 0.005m, and for rotation we use a mean of 0\degree and standard deviation of 0.5\degree. 

Models were trained on 8 TITAN X GPUs for 100,000 episodes using A3C with 32 threads. For each episode, we sample a random training scene and random starting location in the scene for the agent. We use the Adam optimizer with a learning rate of 0.0001.

We train the models on a subset of 50 train scenes and report numbers on 2 of the test-dev scenes. We report metrics on starting locations categorized into \emph{easy}, \emph{medium} and \emph{hard}. If the length of the shortest path from the starting point to the target is among the lowest 20\% path lengths in that scene, it is considered easy. If it is between 20\% and 60\%, it is considered medium, and longer paths are considered as hard. 

\noindent \textbf{Sim-to-Sim} Table~\ref{tab:sim2sim} shows the results of our benchmarked models when trained on simulation and evaluated on the test-dev scenes in simulation. The trivial baselines Random, Instant Done and Blind perform very poorly, indicating that the dataset lacks a bias that is trivial to exploit using these models. The Image only model performs reasonably well, succeeding at over 50\% of easy trajectories but doing very poorly at hard ones. Adding object detection does not help performance. Our analysis in Section~\ref{sec:analysis} shows that object detectors trained in the real world show a drop in performance in simulation, which might be contributing to their ineffectiveness.

\noindent \textbf{Sim-to-Real} Due to the slow motion of the robot, episodes on the robot last as long as 10 minutes. This limits the amount of testing that can be done in the real world. Table~\ref{tab:sim2real} shows the result of the best performing model on Sim-to-Sim, evaluated on a real scene on a subset of starting locations as those reported in Table~\ref{tab:sim2sim}. This shows that there is a sizeable drop in performance for the real robot, especially in SPL. The robot however, does learn to navigate around and explore its environment fairly safely and over 80\% of the trajectories have no collisions with obstacles. This leads to high values for episode lengths in all 3 cases - easy, medium and hard.

\noindent \textbf{Overfit Sim-to-Real} The semantic navigation sim-to-real task in \env\ tests two kinds of generalization: Moving to new scenes and moving from simulation to real. To factor out the former and focus on the latter, we trained a policy on a test-dev scene, (which expectedly led to overfitting on sim) and then deployed this on the robot. Table~\ref{tab:sim2realoverfit} shows these results and demonstrates the upper bound of current models in the real world if they had memorized the test-dev scene perfectly in simulation. The robot does very well on the easy targets, but is affected a lot on hard targets. For all three modes, the SPL is affected tremendously. Appearance and control variations often lead the robot to spaces away from the target, which does not happen in simulation due to overfitting.

\begin{table*}[tp]
 	\centering
 	\setlength{\tabcolsep}{2pt}
 	\begin{tabular}{c|c c c c|c c c c|c c c c|}
    \cline{2-13}
 	   & \multicolumn{4}{|c|}{Easy} & \multicolumn{4}{|c|}{Medium} &
 	    \multicolumn{4}{|c|}{Hard} \\ 
 	   \cline{2-13}
 	   & Success & SPL & Episode & Path & Success & SPL & Episode & Path & Success & SPL & Episode & Path\\
 	   & & & length & length & & & length & length & & & length & length\\
 	    \hline
 	    \multicolumn{1}{|c|}{Random}& 7.58 & 5.32 & 4.36 & 0.34 & 0.00 & 0.00 & 4.27 & 0.30 & 0.00 & 0.00 & 3.06 & 0.19 \\
 	    \multicolumn{1}{|c|}{Instant Done} & 4.55 & 3.79 & 1.00 & 0.00 & 0.00 & 0.00 & 1.00 & 0.00 & 0.00 & 0.00 & 1.00 & 0.00 \\
 	    \multicolumn{1}{|c|}{Blind} & 4.55 & 3.79 & 1.00 & 0.00 & 0.00 & 0.00 & 1.00 & 0.00 & 0.00 & 0.00 & 1.00 & 0.00 \\
 	    \multicolumn{1}{|c|}{Image} & 55.30 & 38.12 & 45.87 & 9.26 & 28.79 & 19.12 & 78.49 & 14.82 & 1.47 & 0.97 & 81.09 & 14.22\\
 	    \multicolumn{1}{|c|}{Image+Detection} & 36.36 & 19.89 & 63.41 & 11.39 & 11.36 & 5.25 & 90.37 & 16.65 & 0.74 & 0.61 & 83.01 & 14.00\\

 	    \hline
 	\end{tabular}
 	\caption{\textbf{Benchmark results for Sim-to-Sim} }
 	\label{tab:sim2sim}
\end{table*} 	

\begin{table*}[h]
 	\centering
 	\setlength{\tabcolsep}{4pt}
 	\begin{tabular}{c|c c c c|c c c c|c c c c|}
    \cline{2-13}
 	   & \multicolumn{4}{|c|}{Easy} & \multicolumn{4}{|c|}{Medium} &
 	    \multicolumn{4}{|c|}{Hard} \\ 
 	   \cline{2-13}
 	   & Success & SPL & Episode & Path & Success & SPL & Episode & Path & Success & SPL & Episode & Path\\
 	   & & & length & length & & & length & length & & & length & length\\
 	    \hline
 	    \multicolumn{1}{|c|}{Image}& 33.33 & 3.53 & 53.16 & 7.18 & 16.66 & 3.70 & 43.83 & 5.33 & 0.00 & 0.00 & 67.83 & 7.00 \\
 	    \hline
 	\end{tabular}
 	\caption{\textbf{Benchmark results for Sim-to-Real} }
 	\label{tab:sim2real}
\end{table*} 	

\begin{table*}[h]
    \vspace{-0.3cm}
 	\centering
 	\setlength{\tabcolsep}{2pt}
 	\begin{tabular}{c|c c c c|c c c c|c c c c|}
    \cline{2-13}
 	   & \multicolumn{4}{|c|}{Easy} & \multicolumn{4}{|c|}{Medium} &
 	    \multicolumn{4}{|c|}{Hard} \\ 
 	   \cline{2-13}
 	   & Success & SPL & Episode & Path & Success & SPL & Episode & Path & Success & SPL & Episode & Path\\
 	   & & & length & length & & & length & length & & & length & length\\
 	    \hline
 	    \multicolumn{1}{|c|}{Image Sim-2-Sim}& 100 & 82.17 & 8.09 & 1.05 & 100 & 86.17 & 27.52 & 4.77 & 94.12 & 83.18 & 42.26 & 7.90\\
 	    \multicolumn{1}{|c|}{Image Sim-2-Real}& 100 & 12.28 & 15.00 & 2.33 & 83.33 & 18.68 & 43.33 & 5.5 & 50 & 28.53 & 30.16 &7.54\\
 	    \hline
 	\end{tabular}
    \vspace{-0.2cm}
 	\caption{\textbf{Benchmark results for Sim-to-Real trained on a single test-dev scene.} }
 	\vspace{-0.3cm}
 	\label{tab:sim2realoverfit}
\end{table*}

 \section{Analysis}
 \label{sec:analysis}
 \vspace{-0.2cm}
 We now dig deeper into the appearance disparities between real and simulation images via t-sne embeddings, object detection results and the output policies for both modalities. We also provide a study showing the effect of changing camera parameters between real and simulation for the transfer problem. Finally, we evaluate using an image translation method for the purposes of domain adaptation.
 
\begin{figure}[tp]
    \centering
    \includegraphics[width=18pc]{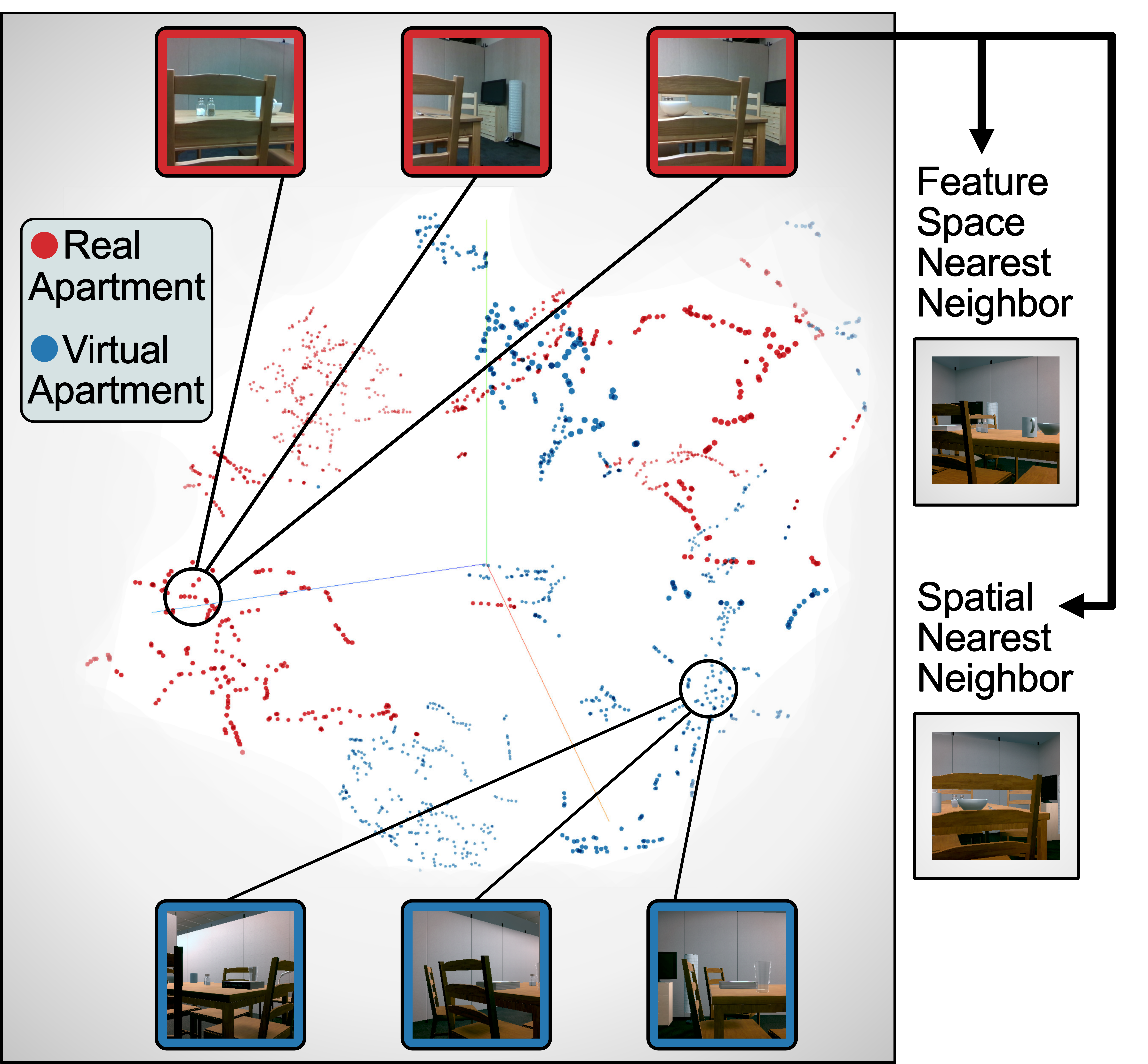}
    \caption{\textbf{Comparison of embeddings for real and synthetic images.} The scatter plot shows a t-SNE visualization of ResNet-18 (pre-trained on ImageNet) features for images from real and simulated apartments. Also shown are the nearest neighbor in feature space and spatial nearest neighbor, which differ slightly in the viewpoint of the agent.}
    \label{fig:tsne} 
    \vspace{-0.3cm}
\end{figure}

\noindent \textbf{Appearance disparities.} 
To the naked eye, images between the real and simulation worlds in \env\ look quite similar. However when we look at embeddings provided by networks, the disparity becomes more visible. We considered 846 images, each from simulation and real collected from the same locations in the scene, passed these images through ResNet-18 to obtain a 512 dim feature vector and then used t-SNE \cite{tsne} to reduce the dimensionality to 3. Figure~\ref{fig:tsne} shows these embeddings. One can clearly see the separation of points into real and simulation clusters. This indicates that embeddings of images in the two modalities are different, which goes towards explaining the drop in performance between simulation and real test-dev scenes (see Table~\ref{tab:sim2sim} and Table~\ref{tab:sim2real}). Figure~\ref{fig:tsne} also shows the nearest neighbor (cosine similarity) simulation image to one of the real images. We found that although nearest neighbors are not far away spatially, they are still slightly different, sometimes having a different view, or a different distance to an obstacle. This might explain why the robot takes different actions in the real world. This analysis indicates that methods of representation learning might need to be revisited, especially when these representations must work across real and simulation modalities.

\begin{figure}[tp]
    \centering
    \includegraphics[width=18pc]{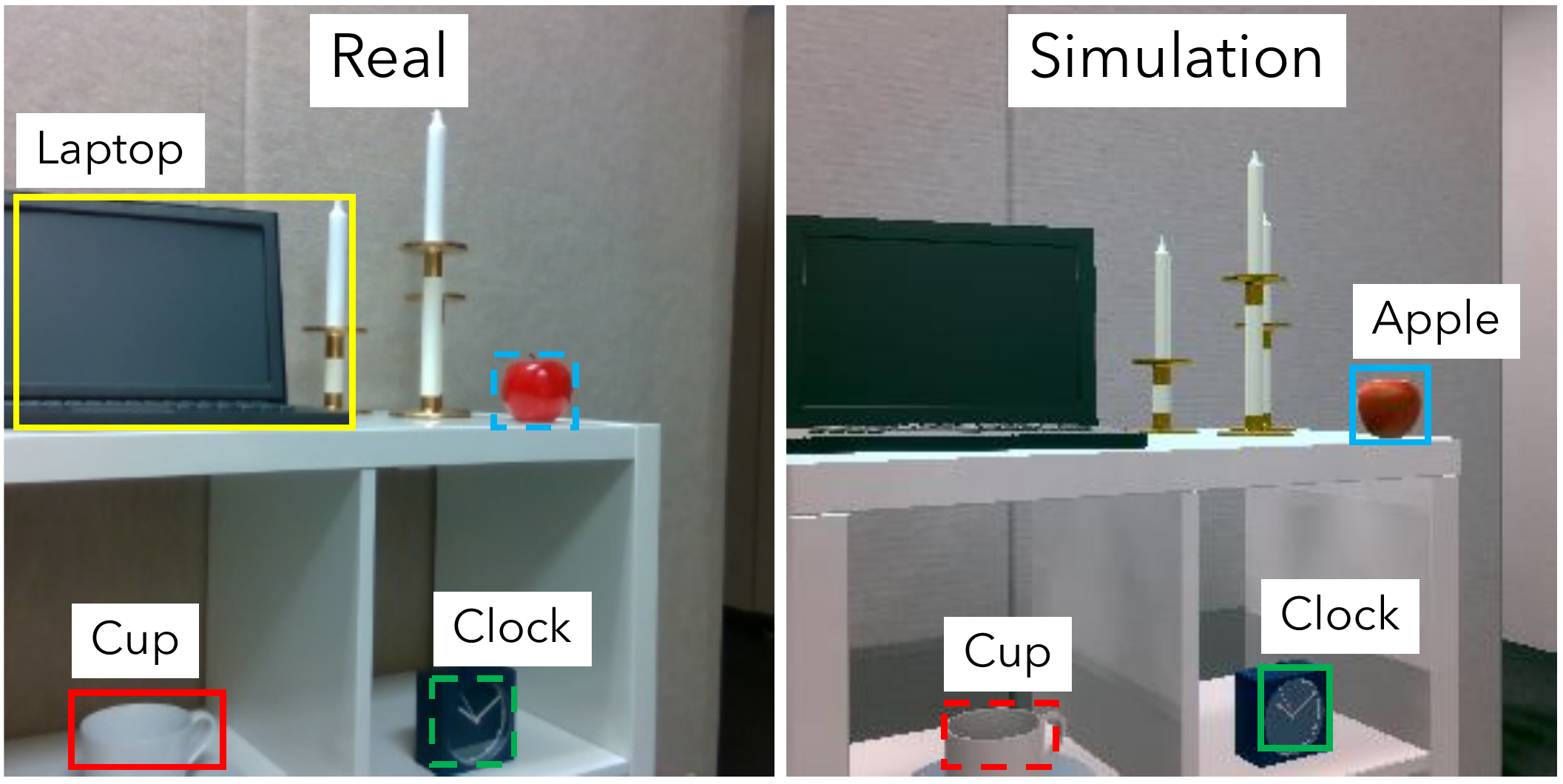}
    \caption{\textbf{Object detection.} Results of object detection in a real and simulated image. Solid lines denote high confidence detections whereas dashed lines denote low confidence detections.}
    \label{fig:obj_det} 
    \vspace{-0.3cm}
\end{figure}

\noindent \textbf{Object detection transfer.} Since we leverage an off the shelf object detector (Faster-RCNN) trained on natural images (MS-COCO), it is imperative to compare the accuracy of this model on images collected in the real and simulated apartments. We collected 761 corresponding images from both environments, ran Faster-RCNN on them and obtained ground truth annotations for 10 object classes (all of which intersected with MS-COCO classes) on Amazon Mechanical Turk. At an Intersection-over-Union (IOU) of 0.5, we obtained a mAP of 0.338 for real images and 0.255 for simulated ones; demonstrating that there is a performance hit going across modalities. Furthermore, detection probabilities tend to differ between the two modalities as demonstrated in Figure~\ref{fig:obj_det}, rendering transfer more challenging for models that exploit these probabilities. Note that this transfer is in the opposite direction (Real to Sim) compared to our current transfer setup, but is revealing nonetheless.

\noindent \textbf{Modifying camera parameters.} 
Since the real and simulation apartments were both designed by us, we were able to model the simulation camera close to the one present on LoCoBot. However, to test the sensitivity of the learnt policy to the camera parameters, we performed an experiment where we trained with a Field Of View (FOV) of 90\degree and tested with an FOV of 42.5\degree. The resulting real world experiments for the Image only model show a huge drop of performance to 16\% for easy, and 0\% for medium and hard. Interestingly, the robot tends to move very little and instead rotate endlessly. We hypothesize that models trained on simulation likely overfit to the image distribution observed at training, and the images captured at a different FOV vary so significantly in feature space, that they invoke unexpected behaviors. Since popular image representation models are trained on images from the internet, they are biased towards the distribution of cameras that people use to take pictures. Cameras on robots are usually quite different. This suggests that we should fine tune representations to match robot cameras and also consider different camera parameters as a camera augmentation step during training in simulation.

\begin{figure}[h]
    \centering
    \includegraphics[width=14pc]{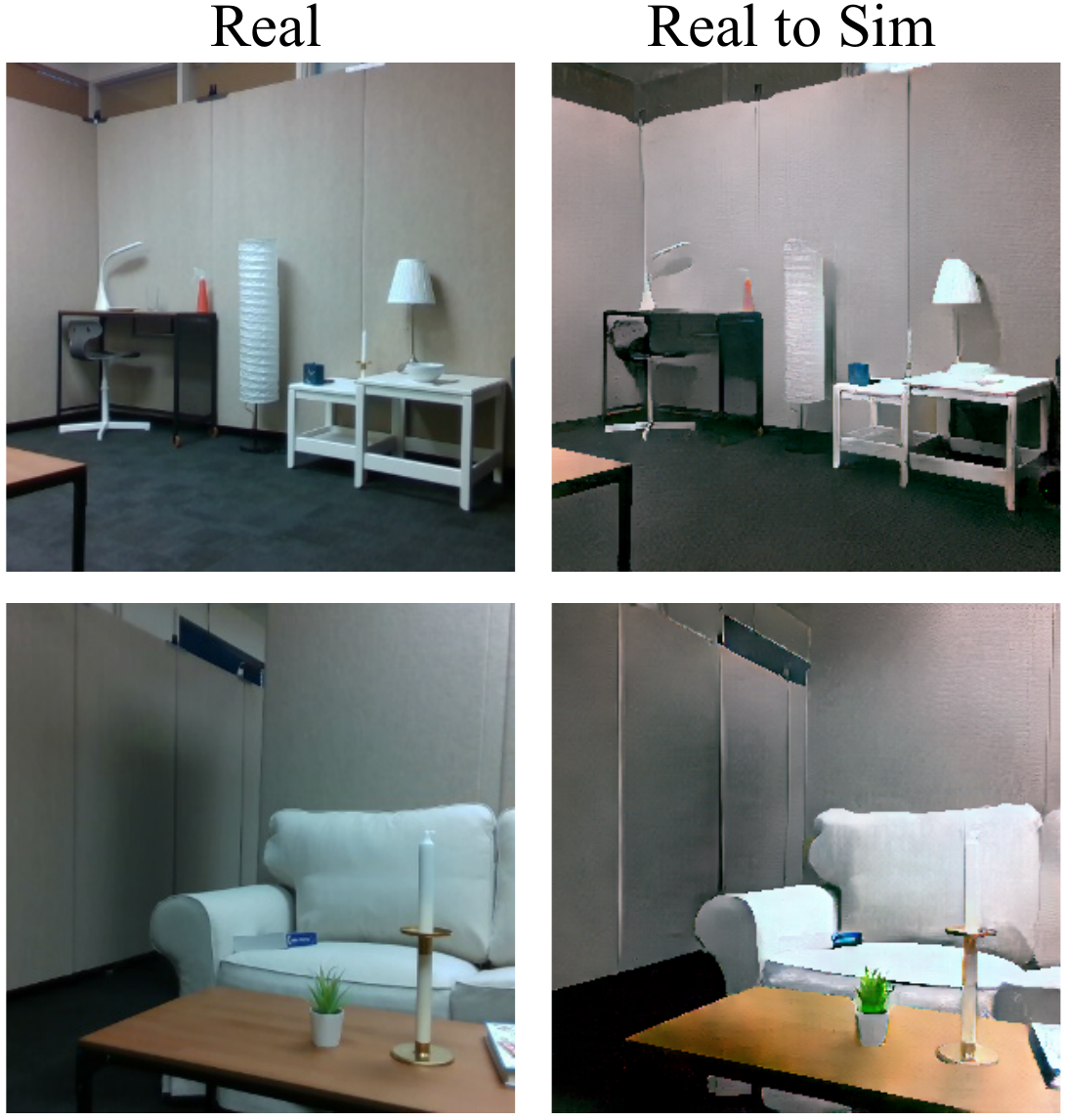}
    \caption{\textbf{Examples of real to simulation transfer.} We use CycleGAN \cite{CycleGAN2017} to translate real images towards simulated ones. The model learns to flatten out the texture and adjust the shadows to look like a simulated image. }
    \label{fig:cyclegan} 
    \vspace{-0.5cm}
\end{figure}

\noindent \textbf{Domain adaptation via image translation.}
Since appearance statistics vary between real and simulation, we experimented with applying an image-to-image translation technique, CycleGAN~\cite{CycleGAN2017} to translate real world images towards simulation images. This would enable us to train in simulation and apply the policy on the real robot while processing the translated images. We needed to use a translation model that could use unpaired image data, since we are unable to obtain paired images with 0 error in the placement of the agent. Paired images were obtained for 3 test-dev scenes to train CycleGAN. The policy (trained in simulation) was run on the robot on the remaining test-dev scene. Interestingly, the CycleGAN model does learn to flatten textures and adjust lighting and shadows as seen in Figure~\ref{fig:cyclegan}. However, the resultant robot performance is very poor and obtains 0\% accuracy. While image translation looks pleasing to the eye, it does introduce spurious errors which hugely affect the image embeddings and thus the resultant policy.
 
\section{Conclusion}
In this paper, we presented \env\ , an open, modular, re-configurable and replicable embodied AI platform with counterparts in simulation and the real world, where researchers across the globe can remotely deploy their models onto physical robots and test their algorithms in the physical world. Our preliminary findings show the performance of models drops significantly when transferring from simulation to real. We hope that \env\ will enable more research towards this important problem.

\noindent\textbf{Acknowledgements.} This work is in part supported by NSF IIS 1652052, IIS 17303166, DARPA N66001-19-2-4031,  67102239  and gifts from  Allen Institute for AI. 

{\small
\bibliographystyle{ieee_fullname}
\bibliography{egbib}
}

\end{document}